\documentclass{article}

\usepackage{PRIMEarxiv}

\usepackage[utf8]{inputenc}
\usepackage[T1]{fontenc}
\usepackage[round,authoryear]{natbib}
\usepackage{hyperref}
\usepackage{url}
\usepackage{booktabs}
\usepackage{amsmath,amssymb,amsthm,mathtools}
\usepackage{nicefrac}
\usepackage{microtype}
\usepackage{fancyhdr}
\usepackage{graphicx}
\usepackage{multirow}
\usepackage[ruled,vlined]{algorithm2e}
\usepackage{xcolor}
\graphicspath{{media/}}
\hypersetup{pdftitle={Pricing Symbolic Rules by What GNNs Leave Unlearned},pdfauthor={Satabarto Sarkar, Prithwineel Paul},colorlinks=true,linkcolor=blue!60!black,citecolor=blue!60!black,urlcolor=magenta!70!black}

\theoremstyle{plain}
\newtheorem{theorem}{Theorem}
\newtheorem{proposition}{Proposition}
\newtheorem{corollary}{Corollary}
\newtheorem{lemma}{Lemma}
\theoremstyle{remark}
\newtheorem{remark}{Remark}

\newcommand{\I}{\mathrm{I}}
\newcommand{\Hh}{\mathrm{H}}
\newcommand{\E}{\mathbb{E}}
\newcommand{\F}{\mathcal{F}}
\newcommand{\WL}{\mathrm{WL}}
\newcommand{\padj}{p_{\mathrm{adj}}}
\newcommand{\lmax}{\ell_{\max}}
\newcommand{\Ct}{\tilde{C}}
\newcommand{\sF}{\sigma_{\F}}
\newcommand{\eps}{\varepsilon}
\newcommand{\ci}[1]{{\scriptsize [#1]}}

\title{Pricing Symbolic Rules by What GNNs Leave Unlearned
\thanks{Accepted at the GraphML Workshop (\emph{Graph Machine Learning: Foundations, Frontiers, and Applications}) at the 18th Asian Conference on Machine Learning (ACML 2026), Melbourne, Australia.}}

\author{
  Satabarto Sarkar \\
  Department of Computer Science and Engineering\\
  University of Engineering and Management\\
  Kolkata 700091, India\\
  \texttt{Satabarto.Sarkar2023@iem.edu.in}
  \And
  Prithwineel Paul \\
  Department of Computer Science and Engineering\\
  University of Engineering and Management\\
  Kolkata 700091, India\\
  \texttt{prithwineel.paul@iem.edu.in}
}

\begin{document}
\maketitle

\begin{abstract}
Symbolic knowledge (motif counts, ring counts, label constraints) is routinely added to graph neural networks (GNNs) and justified by their limited expressivity. We replace this yardstick with an exact one. The largest log-loss reduction a rule $S$ can give a model family is $\I(Y;S\mid\sigma)$ given what the family computes; every realised gain splits exactly into this term plus repaid \emph{neural debt}; and for a trained network with representation $T$ the value is $\I(Y;S\mid T)$, the label information training failed to extract. Rule design thus becomes rule discovery: we score candidates on the network's out-of-fold residual against calibrated permutation nulls with family-wise error control. On ZINC, discovery recovers the hidden ring term $\max(0,\lmax-6)$ of the target in three independent runs ($\padj=0.001$) and beats expert cycle counts; frozen and applied to five unseen datasets it is never significantly worse than the plain network or the expert rule, and on BBBP it significantly beats both.
\end{abstract}


\section{Introduction}
\label{sec:intro}

Neuro-symbolic graph learning augments GNNs with cycle and motif counts \citep{bouritsas2023gsn,bodnar2021cw}, probabilistic label models \citep{qu2019gmnn,huang2021cs} and logic \citep{qu2019pln,zhang2020expressgnn}. The standard argument is expressivity: message passing is at most as strong as 1-WL \citep{xu2019gin,morris2019wl}, which cannot count many substructures \citep{chen2020count}, so the missing structure should be supplied symbolically. This says why rules \emph{can} help, not \emph{how much} a rule is worth to a given network on given data. We prove an \textbf{absorption identity} (Theorem~\ref{thm:absorption}), a \textbf{gain ledger} (Theorem~\ref{thm:ledger}) and a \textbf{learnability decomposition} (Theorem~\ref{thm:learnability}) with a Fano certificate, and turn them into calibrated out-of-fold rule discovery (Section~\ref{sec:method}). Unlike rule extraction, which reads out what a network \emph{has} learned \citep{geng2025logicxgnn,morris2024rgcn}, discovery targets what it \emph{has not}.

\section{Background}
\label{sec:background}

A graph $G$ with label $Y$ is drawn from $P$; predictors are scored by log-loss $L(f)=\E[-\log f(Y\mid G)]$ (nats). A family $\F$ has \emph{computable information} $\sF=\sigma\{h(G):h\in\mathcal{H}_{\F}\}$, where every $f\in\F$ is $f(y\mid G)=g(y,h(G))$. Following \citet{xu2020usable}, $\Hh_{\F}(Y\mid G)=\inf_{f\in\F}L(f)$. The rule-augmented family $\F\oplus S$ adds a branch $\phi(S)$ to the logits, with $\phi\equiv0$ admissible, and the rule's value is
\begin{equation}
\Delta_{\F}(S)=\Hh_{\F}(Y\mid G)-\Hh_{\F\oplus S}(Y\mid G,S)\;\ge0.
\label{eq:value}
\end{equation}
A $t$-layer GIN computes a function of the 1-WL colouring $\WL_t(G)$ and can be injective on the $\WL_t$ classes of any finite sample, so $\sF=\sigma(\WL_t)$ there \citep{xu2019gin}. The \emph{ambiguity} of a rule on a dataset $\{G_i\}_{i=1}^N$ is
\begin{equation}
\alpha_t(s)=\tfrac{1}{N}\bigl|\{i:\exists j,\ \WL_t(G_i)=\WL_t(G_j),\ s(G_i)\ne s(G_j)\}\bigr|,
\label{eq:ambiguity}
\end{equation}
and $\alpha_t(s)=0$ means $s$ is $\WL_t$-measurable on the data. For a trained network, $T=\hat h(G)$ is the input of its readout, and $\sigma(T)\subseteq\sF$.

Throughout, $Y$ and every rule $S$ take values in countable sets with finite entropy. $\sigma$-algebras are identified with the random variables that generate them, and $\Hh(Y\mid\sigma)$ is the conditional entropy given $\sigma$. All information quantities are in nats. The following lemma is used repeatedly.

\begin{lemma}[Gibbs]
\label{lem:gibbs}
Let $\mathcal{A}$ be a $\sigma$-algebra and $q(\cdot\mid\mathcal{A})$ any $\mathcal{A}$-measurable conditional distribution on $\mathcal{Y}$. Then $\E[-\log q(Y\mid\mathcal{A})]\ge\Hh(Y\mid\mathcal{A})$, with equality iff $q=P_{Y\mid\mathcal{A}}$ almost surely.
\end{lemma}
\begin{proof}
$\E[-\log q(Y\mid\mathcal{A})]-\Hh(Y\mid\mathcal{A})=\E\bigl[\sum_y P(y\mid\mathcal{A})\log\frac{P(y\mid\mathcal{A})}{q(y\mid\mathcal{A})}\bigr]=\E\,\mathrm{KL}(P_{Y\mid\mathcal{A}}\,\|\,q)\ge0$, with equality iff the KL divergence vanishes almost surely.
\end{proof}

\section{Theory}
\label{sec:theory}

\subsection{Representational value: the absorption identity}

\begin{theorem}[Absorption identity]
\label{thm:absorption}
Let $\F^\star$ contain all $\sigma$-measurable predictors and $\F^\star\oplus S$ all $\sigma\vee\sigma(S)$-measurable ones. Then
\begin{align}
\Delta^\star(S)&=\I(Y;S\mid\sigma)=\I(Y;S)-\I(S;\sigma)+\I(S;\sigma\mid Y), \label{eq:absorption}\\
\bigl[\I(Y;S)-\I(S;\sigma)\bigr]_+&\le\Delta^\star(S)\le\min\{\Hh(S\mid\sigma),\Hh(Y\mid\sigma)\}, \label{eq:absbounds}
\end{align}
and the upper bound $\Hh(S\mid\sigma)$ exceeds the unclipped lower bound by exactly $\Hh(S\mid Y)$.
\end{theorem}
\begin{proof}
$\F^\star$ contains every $\sigma$-measurable conditional distribution, so by Lemma~\ref{lem:gibbs} $\Hh_{\F^\star}(Y\mid G)=\Hh(Y\mid\sigma)$, attained at $P_{Y\mid\sigma}$. The augmented family contains every $(\sigma\vee\sigma(S))$-measurable conditional distribution whose logits are additive in a $\sigma$-measurable and an $S$-measurable term. When the branch $\phi$ ranges over all functions of $(S,\sigma)$ (the Bayes family), the same lemma gives $\Hh(Y\mid\sigma,S)$. In general it gives at least this value, with equality whenever $\log P(y\mid\sigma,S)-\log P(y\mid\sigma)$ is a function of $(y,S)$ alone. Hence $\Delta^\star=\Hh(Y\mid\sigma)-\Hh(Y\mid\sigma,S)=\I(Y;S\mid\sigma)$, which is the first equality in (\ref{eq:absorption}). The chain rule applied to $\I(S;Y,\sigma)$ in both orders gives
\[
\I(S;\sigma)+\I(S;Y\mid\sigma)=\I(S;Y)+\I(S;\sigma\mid Y),
\]
which rearranges to the second equality in (\ref{eq:absorption}). Since $\I(S;\sigma\mid Y)\ge0$ and $\I(Y;S\mid\sigma)\ge0$, the lower bound in (\ref{eq:absbounds}) holds. For discrete variables, $\I(Y;S\mid\sigma)=\Hh(S\mid\sigma)-\Hh(S\mid Y,\sigma)\le\Hh(S\mid\sigma)$ and symmetrically $\le\Hh(Y\mid\sigma)$, which is the upper bound. Finally,
\[
\Hh(S\mid\sigma)-\bigl[\I(Y;S)-\I(S;\sigma)\bigr]=\Hh(S\mid\sigma)+\I(S;\sigma)-\I(Y;S)=\Hh(S)-\I(Y;S)=\Hh(S\mid Y),
\]
so the gap between the bounds is $\Hh(S\mid Y)$, which is zero iff $S$ is a function of $Y$.
\end{proof}

\paragraph{Additive branches.} The equality case above explains a design choice of rule branches. An additive branch reaches the Bayes gain when the log-likelihood ratio contributed by $S$ does not interact with $\sigma$. Otherwise the realised gain is lower, and the shortfall enters the ledger of Theorem~\ref{thm:ledger} as $\eps_{\F\oplus S}$.

\begin{corollary}[Expressible knowledge is worthless]
\label{cor:worthless}
If $S$ is $\sigma$-measurable, then $\Hh(S\mid\sigma)=0$ and $\Delta^\star(S)=0$. For a $t$-layer GIN, every rule with $\alpha_t(s)=0$ on the data distribution has zero representational value, including rules that 1-WL cannot compute on general graphs.
\end{corollary}
\begin{proof}
If $S$ is $\sigma$-measurable, $\Hh(S\mid\sigma)=0$, and the upper bound in (\ref{eq:absbounds}) gives $\Delta^\star=0$. For a $t$-layer GIN with injective aggregation, $\sF=\sigma(\WL_t)$ on every finite sample \citep{xu2019gin,morris2019wl}. $\alpha_t(s)=0$ means that $s$ is constant on every $\WL_t$ class present, so $s$ is $\WL_t$-measurable on the empirical distribution and its plug-in conditional entropy $\hat\Hh(S\mid\WL_t)$ is zero.
\end{proof}

Hence a rule with $\alpha_t(s)=0$ is worth zero to a Bayes-optimal $t$-layer GIN, even if 1-WL cannot compute it on general graphs.

\subsection{Realised value: the gain ledger}

Real networks are not Bayes-optimal; their \emph{neural debt} is $\eps_{\F}=\Hh_{\F}(Y\mid G)-\Hh(Y\mid\sF)\ge0$.

\begin{theorem}[Gain ledger]
\label{thm:ledger}
For every family $\F$ and rule $S$,
\begin{equation}
\Delta_{\F}(S)=\underbrace{\I(Y;S\mid\sF)}_{\text{representational}}+\underbrace{\eps_{\F}-\eps_{\F\oplus S}}_{\text{debt repaid}},\qquad 0\le\Delta_{\F}(S)\le\I(Y;S\mid\sF)+\eps_{\F}.
\label{eq:ledger}
\end{equation}
Hence a $\sF$-measurable rule can only repay debt, $\Delta_{\F}(S)\le\eps_{\F}$, and if $\F\subseteq\F'$ with $\sigma_{\F'}=\sF$, every nat of debt the larger family repays removes one nat from the ceiling of any such rule.
\end{theorem}
\begin{proof}
By the definitions of the debts, $\Hh_{\F}(Y\mid G)=\Hh(Y\mid\sF)+\eps_{\F}$ and $\Hh_{\F\oplus S}(Y\mid G,S)=\Hh(Y\mid\sF,S)+\eps_{\F\oplus S}$. Subtracting the second from the first and using $\Hh(Y\mid\sF)-\Hh(Y\mid\sF,S)=\I(Y;S\mid\sF)$ yields (\ref{eq:ledger}). Every $f\in\F\oplus S$ is a conditional distribution measurable with respect to $\sF\vee\sigma(S)$, so Lemma~\ref{lem:gibbs} gives $\eps_{\F\oplus S}\ge0$ and therefore the upper bound. Choosing $\phi\equiv0$ shows $\F\subseteq\F\oplus S$, so $\Hh_{\F\oplus S}\le\Hh_{\F}$ and $\Delta_{\F}\ge0$.

For the consequences: if $S$ is $\sF$-measurable, $\I(Y;S\mid\sF)=0$ in (\ref{eq:ledger}), so $\Delta_{\F}(S)=\eps_{\F}-\eps_{\F\oplus S}\le\eps_{\F}$. In particular a test-loss reduction $g$ by such a rule certifies $\eps_{\F}\ge g$ up to estimation error. If $\F\subseteq\F'$, then $\Hh_{\F'}\le\Hh_{\F}$; with $\sigma_{\F'}=\sF$ the Bayes term $\Hh(Y\mid\sF)$ is shared, so $\eps_{\F'}=\eps_{\F}-(\Hh_{\F}-\Hh_{\F'})$, and the ceiling $\eps$ of every $\sF$-measurable rule falls by exactly the debt the larger family repays. If $\eps_{\F}=0$, (\ref{eq:ledger}) reduces to Theorem~\ref{thm:absorption}.
\end{proof}

\subsection{Trained networks: the learnability decomposition}

\begin{theorem}[Learnability decomposition]
\label{thm:learnability}
If $\sigma(T)\subseteq\sF$, the best post-hoc gain of adding $S$ to $T$ is $\Hh(Y\mid T)-\Hh(Y\mid T,S)=\I(Y;S\mid T)$, and
\begin{equation}
\begin{aligned}
\I(Y;S\mid T)&=\I(Y;S\mid\sF)+\underbrace{\I(Y;\sF\mid T)-\I(Y;\sF\mid T,S)}_{\Lambda_T(S):\ \text{lost in training, restored by }S}\\
&\le\I(Y;S\mid\sF)+\bigl[\Hh(Y\mid T)-\Hh(Y\mid\sF)\bigr].
\end{aligned}
\label{eq:learnability}
\end{equation}
\end{theorem}
\begin{proof}
The post-hoc family of all conditional distributions measurable in $(T,S)$ attains $\Hh(Y\mid T,S)$ by Lemma~\ref{lem:gibbs}, and the family measurable in $T$ attains $\Hh(Y\mid T)$, so $\Delta_T(S)=\I(Y;S\mid T)$. By the chain rule,
\[
\I(Y;S,\sF\mid T)=\I(Y;S\mid T)+\I(Y;\sF\mid T,S)=\I(Y;\sF\mid T)+\I(Y;S\mid\sF,T).
\]
Because $\sigma(T)\subseteq\sF$, conditioning on $(\sF,T)$ is conditioning on $\sF$, so $\I(Y;S\mid\sF,T)=\I(Y;S\mid\sF)$. Rearranging gives (\ref{eq:learnability}). Since $\I(Y;\sF\mid T,S)\ge0$, $\Lambda_T(S)\le\I(Y;\sF\mid T)=\Hh(Y\mid T)-\Hh(Y\mid\sF,T)=\Hh(Y\mid T)-\Hh(Y\mid\sF)$. If $S$ is $\sF$-measurable, the first term of (\ref{eq:learnability}) vanishes.
\end{proof}

Expressivity is thus \emph{neither necessary nor sufficient}: a rule $T$ determines is worthless however far beyond 1-WL it lies, and an unlearned $\sF$-measurable rule can be worth $\Lambda_T(S)>0$. Theorems~\ref{thm:ledger} and~\ref{thm:learnability} are two views of one identity. Let $\F_T$ be the family of predictors that read $T$ only. Its debt relative to $\sF$ is $\Hh(Y\mid T)-\Hh(Y\mid\sF)$, exactly the bound in Theorem~\ref{thm:learnability}. The learnability decomposition is therefore the ledger identity for the family ``trained network plus arbitrary post-hoc head'', with the debt made explicit as lost label information. A probe checks which case holds:

\begin{proposition}[Fano certificate]
\label{prop:fano}
For discrete rule columns $S_c$ with alphabets $\mathcal{S}_c$ and any probe $\hat S=\varphi(T)$ with error rates $P_{e,c}$,
\begin{equation}
\I(Y;S\mid T)\le\Hh(S\mid T)\le\Hh(S\mid\hat S)\le\textstyle\sum_c\bigl[h_b(P_{e,c})+P_{e,c}\log(|\mathcal{S}_c|-1)\bigr].
\label{eq:fano}
\end{equation}
\end{proposition}
\begin{proof}
$\I(Y;S\mid T)=\Hh(S\mid T)-\Hh(S\mid T,Y)\le\Hh(S\mid T)$. Since $\hat S=\varphi(T)$, conditioning on $T$ refines conditioning on $\hat S$, so $\Hh(S\mid T)\le\Hh(S\mid\hat S)$. By the chain rule and because conditioning reduces entropy, $\Hh(S\mid\hat S)=\sum_c\Hh(S_c\mid S_{<c},\hat S)\le\sum_c\Hh(S_c\mid\hat S_c)$. Fano's inequality \citep[Thm.~2.10.1]{cover2006} bounds each term by $h_b(P_{e,c})+P_{e,c}\log(|\mathcal{S}_c|-1)$. If $P_{e,c}=0$ for all $c$, the bound is zero.
\end{proof}

\section{Methodology: out-of-fold rule discovery}
\label{sec:method}

Theorem~\ref{thm:learnability} names the rule worth adding: the candidate maximising $\I(Y;S\mid T)$. Discovery therefore \emph{selects} rules from a candidate library: it decides which named, interpretable statistic carries the label information a trained network failed to extract. It does not synthesise new functional forms, and every library used is listed in Section~\ref{sec:setup}. Let $r=Y-\E[Y\mid T]$ be the residual (one-hot minus probabilities for classification) and $\Ct_B=C_B-\E[C_B\mid T]$ the part of a block $C_B$ of standardised candidate columns that $T$ does not predict, with orthonormal basis $Q_B$. The block energy
\begin{equation}
E_B=\tfrac1n\|Q_B^\top r\|_F^2
\label{eq:energy}
\end{equation}
is a local surrogate of value (Proposition~\ref{prop:local}).

\begin{proposition}[Local value of a block]
\label{prop:local}
(a) If $(r,\Ct_B)$ are jointly Gaussian given $T$, then $\I(Y;C_B\mid T)=-\tfrac12\log\bigl(1-E_B^{\mathrm{pop}}/\mathrm{Var}(r)\bigr)$, where $E_B^{\mathrm{pop}}$ is the population variance of the projection of $r$ on $\Ct_B$, and $E_B\to E_B^{\mathrm{pop}}$. (b) Adding $\beta^\top\Ct_B$ to the logit of a binary predictor $p(T)$ changes the expected log-loss by $-\tfrac12g^\top J^{-1}g+o(\|g\|^2)$ at its optimum, with $g=\E[(Y-p)\Ct_B]$ and $J=\E[p(1-p)\Ct_B\Ct_B^\top]$. With whitened $\Ct_B$ and $p(1-p)\approx w$, the reduction is $E_B/(2w)$ to first order. (c) For $K$ classes the exact local reduction is $\tfrac12\mathrm{vec}(G)^\top J^{-1}\mathrm{vec}(G)$ with $G=\E[\Ct_Br^\top]$ and $J=\E[(\mathrm{diag}\,p-pp^\top)\otimes\Ct_B\Ct_B^\top]$; $E_B=\|G\|_F^2$ replaces $J$ by a multiple of the identity, which is exact when the curvature is uncorrelated with $\Ct_B\Ct_B^\top$.
\end{proposition}
\begin{proof}
(a) Let $r=Y-\E[Y\mid T]$ and $\Ct=C_B-\E[C_B\mid T]$ be jointly Gaussian given $T$, with $\Sigma_{rr}=\mathrm{Var}(r)$, $\Sigma_{rc}=\mathrm{Cov}(r,\Ct)$ and $\Sigma_{cc}=\mathrm{Var}(\Ct)$. Then $\I(Y;C_B\mid T)=\I(r;\Ct)=-\tfrac12\log\bigl(1-\Sigma_{rc}\Sigma_{cc}^{-1}\Sigma_{cr}/\Sigma_{rr}\bigr)$. The quantity $\Sigma_{rc}\Sigma_{cc}^{-1}\Sigma_{cr}$ is the variance of the projection of $r$ onto $\Ct$, whose sample analogue with an orthonormal basis $Q_B$ of the centred columns is $\|Q_B^\top r\|^2/n=E_B$. Consistency follows from the law of large numbers for second moments.

(b) Let $\eta=\mathrm{logit}\,p(T)$ and $L(\beta)=\E\bigl[-Y(\eta+\beta^\top\Ct)+\log(1+e^{\eta+\beta^\top\Ct})\bigr]$. Then $\nabla L(0)=-\E[(Y-p)\Ct]=-g$ and $\nabla^2L(0)=\E[p(1-p)\Ct\Ct^\top]=J$. A second-order Taylor expansion gives $L(\beta)=L(0)-\beta^\top g+\tfrac12\beta^\top J\beta+O(\|\beta\|^3)$, which is minimised at $\beta^\star=J^{-1}g+O(\|g\|^2)$ with $L(0)-L(\beta^\star)=\tfrac12g^\top J^{-1}g+o(\|g\|^2)$. If $\E[\Ct\Ct^\top]=I$ and $p(1-p)\approx w$, then $J\approx wI$ and the reduction is $\|g\|^2/(2w)$. With whitened sample rows $z_i=\sqrt n\,Q_{B,i\cdot}$, the plug-in score is $\hat g=\frac1n\sum_ir_iz_i=n^{-1/2}Q_B^\top r$, so $\|\hat g\|^2=E_B$.

(c) For $K$ classes, $r$ is the one-hot label minus the predicted probabilities, and the same argument applied to each logit gives the Frobenius form in (\ref{eq:energy}).
\end{proof}

For Gaussian residuals the energy is therefore an exact monotone transform of value, and for a classifier it is the first-order log-loss reduction. Otherwise $E_B$ only ranks blocks, so significance nominates and validation decides.

\paragraph{Out-of-fold residuals.} In-sample residuals carry a memorisation term that cancels all evidence when $\hat f$ interpolates (Proposition~\ref{prop:oof}). We take $r$ and $T$ from networks trained on the other half of the data (cross-fitting, \citealp{chernozhukov2018dml}).

\begin{proposition}[Out-of-fold versus in-sample evidence]
\label{prop:oof}
Let $\hat f^{(-k)}$ be trained on $D\setminus D_k$. For $i\in D_k$, $\E[(Y_i-\hat f^{(-k)}(G_i))\,c(G_i)\mid\hat f^{(-k)}]=\mathrm{Cov}_P\bigl(Y-\hat f^{(-k)}(G),c(G)\bigr)+\E[Y-\hat f^{(-k)}]\,\E[c]$. For a network $\hat f$ trained on all of $D$ and $\hat f^{(-i)}$ trained without graph $i$,
\begin{equation}
\E\bigl[(Y_i-\hat f(G_i))\,c(G_i)\bigr]=\E\bigl[(Y_i-\hat f^{(-i)}(G_i))\,c(G_i)\bigr]-\underbrace{\E\bigl[(\hat f(G_i)-\hat f^{(-i)}(G_i))\,c(G_i)\bigr]}_{\omega:\ \text{memorisation}}.
\label{eq:memorisation}
\end{equation}
If $\hat f$ interpolates its training labels, the in-sample residual vanishes and the evidence for every candidate collapses.
\end{proposition}
\begin{proof}
For $i\in D_k$, $(G_i,Y_i)$ is independent of $\hat f^{(-k)}$, which is a function of $D\setminus D_k$ and of the training randomness only. Conditioning on $\hat f^{(-k)}$ and writing $e(G,Y)=Y-\hat f^{(-k)}(G)$ gives $\E[e(G_i,Y_i)c(G_i)\mid\hat f^{(-k)}]=\E_P[e\,c]=\mathrm{Cov}_P(e,c)+\E_P[e]\,\E_P[c]$. For the in-sample residual, add and subtract $\hat f^{(-i)}(G_i)$:
\[
Y_i-\hat f(G_i)=\bigl(Y_i-\hat f^{(-i)}(G_i)\bigr)-\bigl(\hat f(G_i)-\hat f^{(-i)}(G_i)\bigr),
\]
then multiply by $c(G_i)$ and take expectations to obtain (\ref{eq:memorisation}). If $\hat f$ interpolates the training labels, $Y_i-\hat f(G_i)=0$ for every $i$, so all in-sample covariances vanish. The leave-one-out residual is unaffected, so the memorisation term $\omega$ exactly cancels the true residual covariance.
\end{proof}

\begin{remark}[Statistical debt]
\label{rem:debt}
The out-of-fold residual measures the debt of a network trained on $n/2$ graphs, part of which disappears at $n$ graphs. A block can therefore be significant for the half-data network and worth little to the full-data network, which is why significance nominates rules and validation loss decides how many to keep. In particular, the half-data networks are used only to compute residuals: every gain we report is measured after retraining on the full training set with the rule added, so a rule that a fully trained network already learns shows no gain there (Section~\ref{sec:oof-evidence}).
\end{remark}

\begin{figure}[t]
\centering
\includegraphics[width=\linewidth]{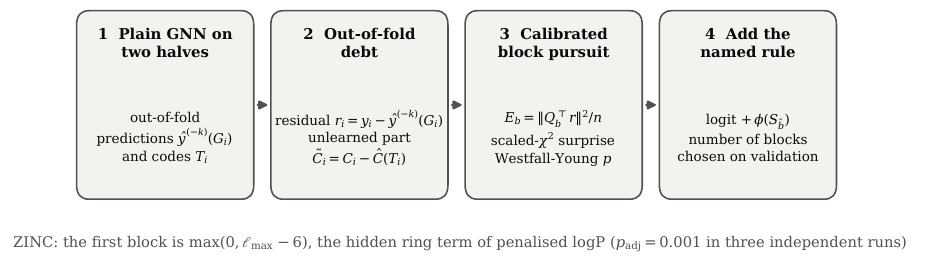}
\caption{Out-of-fold rule discovery. Networks trained on half of the training graphs expose what the architecture fails to generalise; candidate rules are scored by how much of that residual their unlearned part explains, each block against its own permutation null; the winning named rule is added to the network.}
\label{fig:pipeline}
\end{figure}

\paragraph{Calibration and family-wise error.} Blocks differ in size and skew, so we permute the residual rows $\Pi=1000$ times, fit the null energies by a moment-matched scaled $\chi^2$, $a_B\chi^2_{\nu_B}$ with $a_B=\mathrm{Var}/2\mu$ and $\nu_B=2\mu^2/\mathrm{Var}$ \citep{satterthwaite1946,box1954}, and score each block by its surprise $s_B=-\log\Pr(a_B\chi^2_{\nu_B}\ge E_B)$. The pick $\hat B=\arg\max_Bs_B$ receives the Westfall--Young max-$T$ $p$-value \citep{westfall1993}
\begin{equation}
\padj=\frac{1+\#\{\pi:\max_Bs_B^{(\pi)}\ge s_{\hat B}\}}{1+\Pi},
\label{eq:padj}
\end{equation}
which controls the family-wise error rate over all blocks. The residual is projected off the picked span and the search repeats (group OMP, \citealp{pati1993omp}); validation loss picks how many blocks to keep (Algorithm~\ref{alg:discovery}, Figure~\ref{fig:pipeline}).

\paragraph{Scope of the guarantee.} The family-wise error control of (\ref{eq:padj}) is a guarantee for the \emph{first} round: it bounds the probability that the first pick is declared significant when no block carries signal. The $p$-values of later greedy rounds are conditional on the earlier picks and are reported descriptively; the number of blocks kept is decided by validation loss, not by these $p$-values.

\paragraph{Validity of the calibrated max-$T$ test.} Under the complete null that the residual rows are exchangeable with respect to the candidate rows, every permutation $\pi$ of the residual leaves the joint distribution of $(r^\pi,\Ct)$ unchanged. The surprises $s_B$ are fixed monotone transforms of the energies $E_B$, with transforms estimated from the same permutations. The max-$T$ $p$-value (\ref{eq:padj}) is therefore a valid permutation $p$-value for the family-wise null, and it controls the family-wise error rate at any level $\alpha$ \citep{westfall1993}. Strong control holds under subset pivotality. Because the transform is fitted on the permutation null itself, it only rescales the blocks to a common scale and does not affect validity.

\paragraph{Moment-matched scaled chi-square.} If $nE_B/\mathrm{Var}(r)$ were exactly $\chi^2_{d_B}$ (Gaussian residual, fixed basis), its mean and variance would be $d_B$ and $2d_B$. For non-Gaussian residuals and skewed candidate columns, the null of $E_B$ has mean $\mu_B$ and variance $v_B$ that deviate from this. Matching $a\chi^2_\nu$ to them gives $a\nu=\mu_B$ and $2a^2\nu=v_B$, hence $a=v_B/(2\mu_B)$ and $\nu=2\mu_B^2/v_B$ \citep{satterthwaite1946,box1954}. This calibration removes the bias of raw $z$-scores towards single-column blocks, whose nulls are heavy-tailed. We confirmed this on planted signals: both a single planted column and a weak planted ten-column group were recovered, and the pure null gave $\padj\approx0.5$.

\begin{algorithm}[t]
\caption{Out-of-fold rule discovery}
\label{alg:discovery}
\KwIn{training graphs $D$, candidate library $C\in\mathbb{R}^{n\times p}$ with blocks $\mathcal{B}$ (every column and every family), rounds $K$, permutations $\Pi$}
Split $D$ into $D_1,D_2$; for seeds $s=1,2$ train $\hat f_s^{(-k)}$ on $D\setminus D_k$ and keep its best-validation outputs\;
\For{$i\in D_k$}{
  $\hat y_i\leftarrow\frac12\sum_s\hat f_s^{(-k)}(G_i)$,\ \ $T_i\leftarrow\hat h^{(-k)}(G_i)$,\ \ $r_i\leftarrow y_i-\hat y_i$\;
}
\For{$k=1,2$}{
  $\Ct[D_k]\leftarrow C[D_k]-\widehat{C}(T)[D_k]$ \tcp*{cross-fitted MLP probe within fold}
}
$\mathcal{P}\leftarrow\emptyset$,\ \ $r\leftarrow r-\bar r$\;
\For{$j=1,\dots,K$}{
  \lFor{$B\in\mathcal{B}$}{$Q_B\leftarrow\mathrm{orth}(\Ct_B)$,\ \ $E_B\leftarrow\|Q_B^\top r\|_F^2/n$}
  permute rows of $r$ $\Pi$ times, compute null energies, fit $a_B\chi^2_{\nu_B}$, surprises $s_B$\;
  $\hat B\leftarrow\arg\max_{B\notin\mathcal{P}}s_B$,\ \ $\padj$ by (\ref{eq:padj}),\ \ $\mathcal{P}\leftarrow\mathcal{P}\cup\{\hat B\}$\;
  $r\leftarrow r-\Pi_{\mathcal{P}}r$ \tcp*{project off the picked span}
}
Train the network with rule branches for the nested sets $\mathcal{P}_1\subset\dots\subset\mathcal{P}_K$; keep the set with the best validation loss\;
\end{algorithm}

\paragraph{Cost.} A discovery round costs $O(\Pi\,n\,D\,K)$ for $n$ graphs, total block dimension $D$ and $K$ residual columns, on top of training four half-data networks. This took seconds to minutes on one GPU for a few hundred candidates.

\section{Experimental setup}
\label{sec:setup}

\paragraph{Data.} Data are real only: TU benchmarks \citep{morris2020tudataset}, ZINC-12k \citep{dwivedi2023benchmarking}, scaffold-split OGB molecules \citep{hu2020ogb} and eleven node-classification graphs (Table~\ref{tab:datasets}). Sources: TU benchmarks, \url{https://chrsmrrs.github.io/datasets/}; OGB, \url{https://ogb.stanford.edu/docs/graphprop/}; ZINC-12k, \url{https://github.com/graphdeeplearning/benchmarking-gnns}; node-level graphs, \url{https://github.com/kimiyoung/planetoid}, \url{https://github.com/shchur/gnn-benchmark}, \url{https://github.com/graphdml-uiuc-jlu/geom-gcn} and \url{https://github.com/yandex-research/heterophilous-graphs}. ZINC, molhiv and COLLAB served for development; five datasets were held out.

\begin{table}[t]
\centering
\caption{Datasets. Splits: CV = stratified 10-fold cross-validation (WL, absorption and ledger experiments); 80/10/10 = fixed stratified split (discovery experiments); scaffold = OGB scaffold split; ZINC = the standard 10k/1k/1k split.}
\label{tab:datasets}
\small
\begin{tabular}{@{}llrcl@{}}
\toprule
Dataset & task & graphs & classes & splits used\\
\midrule
MUTAG, PTC & binary classification & 188, 344 & 2 & CV\\
PROTEINS & binary classification & 1113 & 2 & CV, 80/10/10, random/size\\
NCI1 & binary classification & 4110 & 2 & CV, random/size\\
IMDB-BINARY & binary classification & 1000 & 2 & CV, 80/10/10, random/size\\
IMDB-MULTI & 3-class classification & 1500 & 3 & CV\\
COLLAB & 3-class classification & 5000 & 3 & CV, 80/10/10\\
REDDIT-BINARY & binary classification & 2000 & 2 & 80/10/10\\
ZINC-12k & regression (MAE) & 12000 & -- & standard\\
ogbg-molhiv & binary (ROC-AUC) & 41127 & 2 & scaffold, random\\
ogbg-molbace & binary (ROC-AUC) & 1513 & 2 & scaffold, random\\
ogbg-molbbbp & binary (ROC-AUC) & 2039 & 2 & scaffold, random\\
\midrule
\multicolumn{5}{@{}p{0.93\linewidth}@{}}{\emph{Node level:} Cora, CiteSeer, PubMed, Amazon-Photo, Amazon-Computers, Actor, Roman-empire, Amazon-ratings, Minesweeper, Tolokers, Questions; label rates 5\% and 50\%; 5 seeds.}\\
\bottomrule
\end{tabular}
\end{table}

\paragraph{Rule libraries.} The \emph{classical library} has eleven rules: graph size, degree histogram, node-type histogram, closed-walk counts, cyclomatic number, triangle counts, cycle counts of length 3--8, ring-atom counts, Laplacian spectral moments, bond-type histogram and a Gaussian-noise placebo. The \emph{discovery library} adds, for molecules, chordless-ring counts by size (3--12), ring summaries (number of rings, $\lmax$, $\max(0,\lmax-6)$, rings with more than six atoms), fusion and spiro counts and the 150 most frequent ring compositions; for social graphs, 26 statistics of size, density, degree shape, triangles and clustering, cores and cliques, components, spectrum and assortativity. The classical library is included as distractors. The same discovery library is used for every molecular dataset; the term $\max(0,\lmax-6)$ is one of its ring summaries, among about 300 ring statistics.

\paragraph{Networks.} Models are four-layer GIN \citep{xu2019gin} or GINE with all OGB features \citep{hu2020pretrain}; rules enter as an MLP branch on the logits. GIN updates node states as $h_v'=\mathrm{MLP}\bigl((1+\epsilon)h_v+\sum_{u\in N(v)}h_u\bigr)$ with a two-layer MLP and batch normalisation. GINE adds a learned embedding of the bond features to each message before the ReLU. On OGB data, the nine atom and three bond features each have their own embedding table, and the embeddings are summed, as in the OGB AtomEncoder and BondEncoder. The representation $T$ is the concatenation of the sum-pooled node states of all layers, including the input layer, so $\dim T=5\times$hidden. The readout is batch normalisation followed by a two-layer MLP. A rule branch maps the standardised rule columns through a two-layer MLP to the output dimension, and its output is added to the logits. For efficiency, all members of one experiment (seeds $\times$ rule sets $\times$ fold networks) are trained jointly as one batched network. Each member has its own weights and batch-normalisation statistics, all members share mini-batches, and each keeps the outputs of its own best validation epoch. Table~\ref{tab:hparams} lists the hyperparameters.

\begin{table}[t]
\centering
\caption{Hyperparameters. All models use four layers and Adam.}
\label{tab:hparams}
\footnotesize
\setlength{\tabcolsep}{4pt}
\resizebox{\linewidth}{!}{\begin{tabular}{@{}llcccccc@{}}
\toprule
Experiment & model & hidden & lr & schedule & dropout & epochs & batch\\
\midrule
WL/absorption/ledger (TU, 10-fold) & GIN, DeepSets & 64 & $10^{-2}$ & patience 100 & 0.5 & 350 & --\\
Classical rules on ZINC/NCI1/PROTEINS & GIN, GINE & 128 & $10^{-3}$ & cosine & 0 & 200 & 128\\
Discovery, ZINC & GINE & 128 & $10^{-3}$ & cosine & 0 & 200 & 128\\
Discovery, molhiv & GINE (OGB feat.) & 256 & $10^{-3}$ & cosine & 0.5 & 80 & 128\\
Discovery, molbace/molbbbp & GINE (OGB feat.) & 256 & $10^{-3}$ & cosine & 0.5 & 100 & 32\\
Discovery, COLLAB/TU & GIN (degree tags) & 64 & $10^{-3}$ & cosine & 0 & 150 & 32\\
Random-split absorption, OGB & GINE (OGB feat.) & 256 (128 molhiv) & $10^{-3}$ & cosine & 0.5 & 100 (60) & 32 (128)\\
Random-split absorption, TU & GIN & 64 & $10^{-3}$ & cosine & 0 & 150 & 32\\
\bottomrule
\end{tabular}}
\end{table}

\paragraph{Probes and $R^2_T$.} The probe that predicts a rule (or a candidate) from $T$ is a two-layer MLP with 64 hidden units, trained with Adam (learning rate $3\times10^{-3}$, weight decay $10^{-4}$) for 300 steps with early stopping on a held-out fifth. It is cross-fitted over two folds, so every prediction is out-of-sample. $R^2_T=1-\E\|S-\hat S(T)\|^2/\mathrm{Var}(S)$ is averaged over the rule's columns.

\paragraph{Discovery.} Out-of-fold residuals come from two seeds $\times$ two folds of networks trained on half of the training graphs. For every training graph, the residual uses the average prediction of the two networks that did not see it, and the representation comes from the first of them. Candidate columns are signed-log transformed, standardised on training graphs and clipped to $[-6,6]$. Ring patterns enter the pool if they occur in at least 20 training graphs (at most 150 patterns), and chordless rings are enumerated up to length 12. Blocks are every single column plus every family. The families are: each classical rule, ring sizes, ring summary, ring fusion, ring atom patterns, ring bond patterns, size, degree shape, triangles and clustering, cores and cliques, components and spectrum. We use $\Pi=1000$ residual permutations per round and three rounds. Rule sets $\mathrm{blk}_1\subset\mathrm{blk}_2\subset\mathrm{blk}_3$ are retrained on the full training set with the same seeds as the plain network, and the set with the lowest mean validation loss is reported. Controls are the expert rule, a size-matched random set of non-picked candidate columns, and the whole pool.

\paragraph{Statistics.} Every comparison is paired by seed. Intervals are 95\% percentile bootstrap intervals of the mean paired difference (4000 resamples). Classification log-losses are per graph in nats. OGB atom types are atomic number minus one, so in ring patterns 5 is C, 6 is N and 7 is O; bond types are 0 single, 1 double, 2 triple and 3 aromatic.

\section{Results}
\label{sec:results}

\subsection{Expressible knowledge is absorbed}

\begin{figure}[t]
\centering
\includegraphics[width=0.62\linewidth]{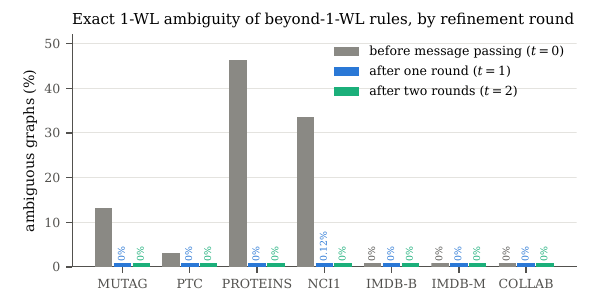}
\caption{Exact 1-WL ambiguity of beyond-1-WL rules, maximum over six rules (Table~\ref{tab:wl}). One round of colour refinement removes the ambiguity almost entirely, and two rounds remove it completely.}
\label{fig:wl}
\end{figure}

Computing $\alpha_t$ exactly on every graph of seven TU benchmarks (Table~\ref{tab:wl}, Figure~\ref{fig:wl}), up to 46\% of graphs are ambiguous before message passing (PROTEINS), at most 0.12\% after one round (NCI1) and none after two. By Theorem~\ref{thm:absorption}, six beyond-1-WL rules are worth zero to a GIN of depth two; their gain to a graph-blind DeepSets model (up to $+11.3$ points) largely vanishes under GIN. GNNs absorb a median 99\% of a belief-propagation layer's value on eleven node-classification graphs, and where GIN still gains, an MLP readout repays 80--83\% of it (COLLAB, NCI1), as the ledger predicts.

\begin{table}[t]
\centering
\caption{Exact 1-WL ambiguity $\alpha_t$ (\%, maximum over six beyond-1-WL rules) and test-accuracy gain of adding those rules (10-fold CV).}
\label{tab:wl}
\small
\begin{tabular}{@{}lccccccc@{}}
\toprule
 & MUTAG & PTC & PROTEINS & NCI1 & IMDB-B & IMDB-M & COLLAB\\
\midrule
$\alpha_0$ & 13.3 & 3.2 & 46.4 & 33.5 & 0 & 0 & 0\\
$\alpha_1$ / $\alpha_{\ge2}$ & 0 / 0 & 0 / 0 & 0 / 0 & 0.12 / 0 & 0 / 0 & 0 / 0 & 0 / 0\\
gain, DeepSets & $+11.3$ & $+2.0$ & $+4.6$ & $+3.1$ & $+1.6$ & $-0.5$ & $+1.9$\\
gain, GIN & $-1.6$ & $-3.8$ & $+2.4$ & $+0.8$ & $-0.3$ & $+0.3$ & $+2.1$\\
\bottomrule
\end{tabular}
\end{table}

\subsection{The gain ledger on graph-level benchmarks}

\begin{figure}[t]
\centering
\includegraphics[width=0.62\linewidth]{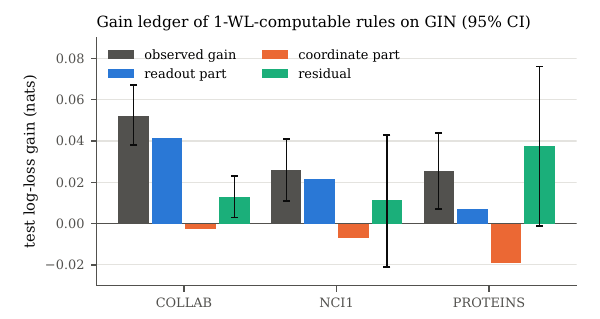}
\caption{Gain ledger of 1-WL-computable rules on GIN (Table~\ref{tab:ledger}). The observed gain decomposes into the part repaid by an MLP readout, the part repaid by log-size coordinates, and the residual. Error bars are the 95\% intervals of the observed gain and the residual.}
\label{fig:ledger}
\end{figure}

For 1-WL-computable rules on GIN, the ledger of Theorem~\ref{thm:ledger} splits the observed gain into the part repaid by an MLP readout, the part repaid by log-size coordinates and a residual (Table~\ref{tab:ledger}, Figure~\ref{fig:ledger}). Most of the gain on COLLAB and NCI1 is debt that a better readout also repays (Table~\ref{tab:readouts}). For beyond-1-WL rules, the ledger gives an observed gain of $+0.020$ nats on COLLAB, with a residual of $+0.012$ \ci{$+0.003,+0.021$} after the readout and coordinate parts. On PROTEINS the observed gain is $+0.023$, with a residual of $+0.032$ \ci{$-0.002,+0.066$}. On every other dataset the observed gain is not significant. The shuffled-rule placebo is significantly \emph{negative} on COLLAB ($-0.052$), IMDB-B ($-0.047$) and IMDB-M ($-0.031$) and never significantly positive, because an extra branch that carries no information can only overfit.

\begin{table}[t]
\centering
\caption{Gain ledger for 1-WL-computable rules on GIN (10-fold CV, test log-loss gain in nats, 95\% CI). The readout and coordinate parts are the reductions in rule gain when the linear readout is replaced by an MLP readout and when log-size coordinates are added.}
\label{tab:ledger}
\small
\begin{tabular}{@{}lcccc@{}}
\toprule
Dataset & observed gain & readout part & coordinate part & residual\\
\midrule
COLLAB & $+0.052$ \ci{$+0.038,+0.067$} & $+0.042$ & $-0.003$ & $+0.013$ \ci{$+0.003,+0.023$}\\
NCI1 & $+0.026$ \ci{$+0.011,+0.041$} & $+0.022$ & $-0.007$ & $+0.011$ \ci{$-0.021,+0.043$}\\
PROTEINS & $+0.026$ \ci{$+0.007,+0.044$} & $+0.007$ & $-0.019$ & $+0.038$ \ci{$-0.001,+0.076$}\\
\bottomrule
\end{tabular}
\end{table}

\begin{table}[t]
\centering
\caption{Plain GIN with three readouts (10-fold CV): test log-loss (nats) and accuracy. The MLP readout repays most of the debt that rules otherwise repay (Table~\ref{tab:ledger}).}
\label{tab:readouts}
\small
\begin{tabular}{@{}lcccccc@{}}
\toprule
 & \multicolumn{3}{c}{log-loss} & \multicolumn{3}{c}{accuracy}\\
\cmidrule(lr){2-4}\cmidrule(lr){5-7}
Dataset & linear & MLP & log-MLP & linear & MLP & log-MLP\\
\midrule
COLLAB & 0.518 & 0.468 & 0.470 & 0.751 & 0.790 & 0.792\\
NCI1 & 0.484 & 0.479 & 0.474 & 0.783 & 0.788 & 0.783\\
PROTEINS & 0.574 & 0.550 & 0.575 & 0.735 & 0.731 & 0.704\\
IMDB-B & 0.516 & 0.533 & 0.529 & 0.726 & 0.724 & 0.721\\
\bottomrule
\end{tabular}
\end{table}

\subsection{Learnability, not expressivity, decides}

\begin{figure}[t]
\centering
\includegraphics[width=\linewidth]{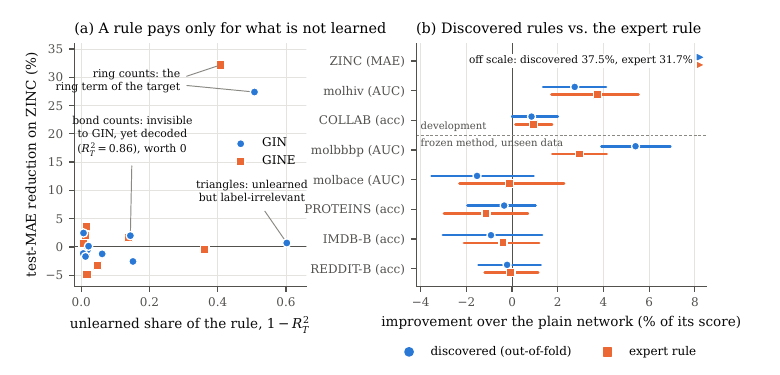}
\caption{(a) ZINC: a rule's gain tracks its unlearned share $1-R^2_T$, not 1-WL expressivity. (b) Discovered (blue) and expert (orange) rules vs.\ the plain network, 95\% intervals; rows below the dashed line were held out.}
\label{fig:main}
\end{figure}

On ZINC (Figure~\ref{fig:main}a, Table~\ref{tab:classical}) bond-type counts are invisible to GIN, yet it decodes them from atoms ($R^2_T=0.86$) and they are worth nothing. Ring counts are the least learned rules ($R^2_T=0.49/0.59$) and the only ones that pay, cutting MAE by 27\% (GIN) and 32\% (GINE). Triangles are unlearned ($R^2_T=0.40$) but worthless: large $\Hh(S\mid T)$ is necessary, not sufficient. On NCI1 and PROTEINS (GIN, same protocol), the largest single-rule gains in test log-loss are $+0.021$ (cyclomatic number on NCI1, $R^2_T=0.95$) and $+0.037$ (type histogram on PROTEINS, $R^2_T=0.71$). Both are learnable rules that repay debt in the sense of Theorem~\ref{thm:ledger}. The Fano bound of Proposition~\ref{prop:fano} held for all 14 rules for which it is defined, at 0.26--32 nats.

\begin{table}[t]
\centering
\caption{Classical rules added to GIN and GINE on ZINC (3 seeds). Gain: reduction in test MAE (plain GIN 0.2742, GINE 0.1918) with 95\% interval. $R^2_T$: share of the rule decoded from the plain network's representation. Probe pick: the two rules with the largest residual score, chosen before any rule model was trained.}
\label{tab:classical}
\small
\begin{tabular}{@{}lcccc@{}}
\toprule
 & \multicolumn{2}{c}{GIN} & \multicolumn{2}{c}{GINE}\\
\cmidrule(lr){2-3}\cmidrule(lr){4-5}
Rule & $R^2_T$ & gain \ci{95\% CI} & $R^2_T$ & gain \ci{95\% CI}\\
\midrule
cycles (3--8) & 0.49 & $+0.0751$ \ci{$+0.0643,+0.0900$} & 0.59 & $+0.0617$ \ci{$+0.0565,+0.0667$}\\
triangles & 0.40 & $+0.0019$ \ci{$-0.0206,+0.0168$} & 0.64 & $-0.0008$ \ci{$-0.0085,+0.0052$}\\
spectrum & 0.85 & $-0.0070$ \ci{$-0.0286,+0.0083$} & 0.86 & $+0.0032$ \ci{$-0.0044,+0.0114$}\\
bond histogram & 0.86 & $+0.0054$ \ci{$-0.0069,+0.0226$} & 0.98 & $-0.0014$ \ci{$-0.0041,+0.0039$}\\
ring atoms & 0.94 & $-0.0034$ \ci{$-0.0196,+0.0055$} & 0.95 & $-0.0063$ \ci{$-0.0165,-0.0010$}\\
degree histogram & 0.98 & $+0.0004$ \ci{$-0.0179,+0.0163$} & 0.98 & $-0.0093$ \ci{$-0.0106,-0.0070$}\\
cyclomatic number & 0.98 & $+0.0068$ \ci{$-0.0094,+0.0198$} & 0.98 & $+0.0068$ \ci{$-0.0022,+0.0167$}\\
type histogram & 0.99 & $-0.0046$ \ci{$-0.0123,+0.0086$} & 0.99 & $+0.0040$ \ci{$-0.0012,+0.0092$}\\
closed walks & 0.99 & $+0.0067$ \ci{$-0.0094,+0.0163$} & 0.99 & $+0.0011$ \ci{$-0.0044,+0.0041$}\\
size & 0.99 & $-0.0032$ \ci{$-0.0150,+0.0050$} & 0.99 & $+0.0042$ \ci{$-0.0010,+0.0071$}\\
noise (placebo) & -- & $-0.0003$ \ci{$-0.0018,+0.0027$} & -- & $-0.0018$ \ci{$-0.0083,+0.0103$}\\
\midrule
probe pick (top 2) & & $+0.0755$ \ci{$+0.0627,+0.0820$} & & $+0.0608$ \ci{$+0.0536,+0.0648$}\\
all rules & & $+0.0745$ \ci{$+0.0614,+0.0872$} & & $+0.0523$ \ci{$+0.0500,+0.0553$}\\
\bottomrule
\end{tabular}
\end{table}

\subsection{Discovery recovers the hidden term of the target}

ZINC's penalised logP subtracts the ring penalty $\max(0,\lmax-6)$ from Crippen logP \citep{gomezbombarelli2018,jin2018jtvae}; atom-additive logP is easy for message passing, the longest-cycle term is not. Among 300 ring statistics plus the classical library, the first pick is exactly $\max(0,\lmax-6)$ for GIN and GINE in three independent runs ($\padj=0.001$, the minimum for $\Pi=1000$). The method selects this term from the library; what the result shows is that, among hundreds of plausible ring statistics and distractors, the out-of-fold residual singles out the one term of the target that message passing leaves unlearned. The validation-selected set reaches MAE 0.1196 against 0.1306 for expert cycle counts ($+0.011$ \ci{$0.009,0.012$}). On molhiv it adds $+0.021$ \ci{$+0.010,+0.031$} ROC-AUC (Tables~\ref{tab:discovery} and~\ref{tab:dev}).

\begin{figure}[t]
\centering
\includegraphics[width=0.85\linewidth]{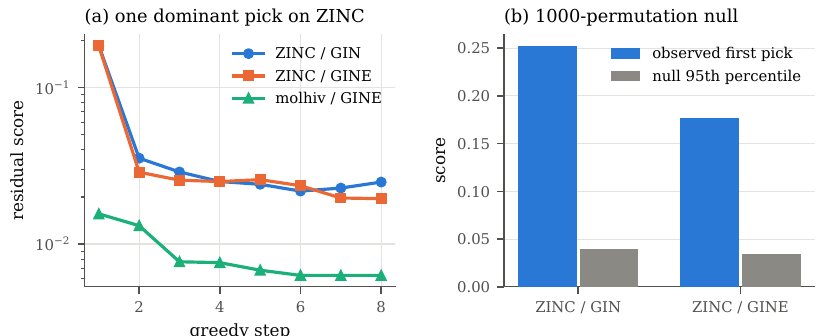}
\caption{(a) Greedy residual scores of the first discovery run (log scale). On ZINC the first pick, the ring term of the target, dominates every later pick by a factor of about five. (b) Fresh-seed run: the first pick against the 95th percentile of its 1000-permutation null.}
\label{fig:scores}
\end{figure}

On ZINC the first pick dominates every later pick by a factor of about five, and it lies far above the 95th percentile of its 1000-permutation null (Figure~\ref{fig:scores}). Earlier discovery runs on ZINC confirm the result. The first discovery run used single columns, in-sample residuals and $k\in\{1,3,8\}$ picks chosen by validation. On GIN it reached MAE 0.1829 (plain 0.2673, cycles 0.1951, whole pool 0.1653, random 8 columns 0.2057). On GINE it reached 0.1186 (plain 0.1938, cycles 0.1333, whole pool 0.1277, random 0.1118). The fresh-seed certification run reached 0.1779 on GIN (plain 0.2722, cycles 0.1965, whole pool 0.1659) and 0.1173 on GINE (plain 0.1885, cycles 0.1322, whole pool 0.1269, random 0.1147). Discovery beat the expert cycle counts in all four cases, by between 0.012 and 0.019 MAE with intervals above zero. Earlier molhiv runs fed the network only atom and bond type and are superseded by the full-feature protocol of Table~\ref{tab:discovery}.

\begin{table}[t]
\centering
\caption{Out-of-fold discovery with the frozen method. Expert: cycle counts (molecules, PROTEINS) or graph size (social graphs). Gains are paired by seed with 95\% bootstrap intervals; for MAE a positive gain means lower error. First block: the first pick and its Westfall--Young $\padj$.}
\label{tab:discovery}
\footnotesize
\setlength{\tabcolsep}{3.5pt}
\resizebox{\linewidth}{!}{\begin{tabular}{@{}lcccccccl@{}}
\toprule
Dataset (metric) & seeds & plain & expert & discovered & gain vs.\ plain & gain vs.\ expert & first block ($\padj$)\\
\midrule
\multicolumn{8}{@{}l}{\emph{development}}\\
ZINC (MAE$\downarrow$) & 3 & 0.1911 & 0.1306 & \textbf{0.1196} & $+0.072$ \ci{.065, .077} & $+0.011$ \ci{.009, .012} & ring term (.001)\\
molhiv (AUC) & 8 & 0.7588 & 0.7871 & 0.7796 & $+0.021$ \ci{.010, .031} & $-0.008$ \ci{$-$.023, .010} & 12-ring, arom.\ (.001)\\
COLLAB (acc) & 5 & 0.7996 & 0.8072 & 0.8064 & $+0.007$ \ci{.000, .016} & $-0.001$ \ci{$-$.008, .006} & clustering (.001)\\
\midrule
\multicolumn{8}{@{}l}{\emph{held out, frozen method}}\\
molbbbp (AUC) & 10 & 0.6421 & 0.6610 & \textbf{0.6768} & $+0.035$ \ci{.025, .045} & $+0.016$ \ci{.002, .030} & C$_5$O 6-ring (.046)\\
molbace (AUC) & 10 & 0.8084 & 0.8074 & 0.7959 & $-0.012$ \ci{$-$.029, .008} & $-0.011$ \ci{$-$.024, .000} & walks (.006)\\
PROTEINS (acc) & 10 & 0.7786 & 0.7696 & 0.7759 & $-0.003$ \ci{$-$.015, .008} & $+0.006$ \ci{$-$.005, .017} & assortativity (.105)\\
IMDB-B (acc) & 10 & 0.7590 & 0.7560 & 0.7520 & $-0.007$ \ci{$-$.023, .010} & $-0.004$ \ci{$-$.017, .008} & type hist.\ (.995)\\
REDDIT-B (acc) & 10 & 0.9175 & 0.9170 & 0.9155 & $-0.002$ \ci{$-$.014, .012} & $-0.002$ \ci{$-$.015, .011} & mean degree (.001)\\
\bottomrule
\end{tabular}}
\end{table}

\begin{table}[t]
\centering
\caption{Development datasets with the frozen out-of-fold method: every configuration, paired gain over the plain network with 95\% interval (MAE: positive $=$ lower error). Bold: validation-selected. Blocks in order of discovery. ZINC: $\max(0,\lmax-6)$; 10-ring count; closed walks. molhiv: 12-ring with seven aromatic bonds; 7-ring with two aromatic bonds; type histogram. COLLAB: triangles/clustering family; 90th percentile of degree.}
\label{tab:dev}
\footnotesize
\setlength{\tabcolsep}{3.5pt}
\resizebox{\linewidth}{!}{\begin{tabular}{@{}lcccccc@{}}
\toprule
 & \multicolumn{2}{c}{ZINC (MAE, 3 seeds)} & \multicolumn{2}{c}{molhiv (AUC, 8 seeds)} & \multicolumn{2}{c}{COLLAB (acc, 5 seeds)}\\
\cmidrule(lr){2-3}\cmidrule(lr){4-5}\cmidrule(lr){6-7}
Config & score & gain & score & gain & score & gain\\
\midrule
plain & 0.1911 & -- & 0.7588 & -- & 0.7996 & --\\
expert rule & 0.1306 & $+0.061$ \ci{.056, .065} & 0.7871 & $+0.028$ \ci{.013, .042} & 0.8072 & $+0.008$ \ci{.001, .014}\\
blk$_1$ & 0.1181 & $+0.073$ \ci{.066, .079} & 0.7741 & $+0.015$ \ci{.007, .025} & \textbf{0.8064} & $+0.007$ \ci{.000, .016}\\
blk$_2$ & \textbf{0.1196} & $+0.072$ \ci{.065, .077} & \textbf{0.7796} & $+0.021$ \ci{.010, .031} & 0.8092 & $+0.010$ \ci{.002, .024}\\
blk$_3$ & 0.1215 & $+0.070$ \ci{.066, .076} & 0.7783 & $+0.020$ \ci{.005, .033} & 0.8052 & $+0.006$ \ci{$-$.003, .015}\\
random (size-matched) & 0.1814 & $+0.010$ \ci{.005, .013} & 0.7742 & $+0.015$ \ci{.008, .023} & 0.7968 & $-0.003$ \ci{$-$.016, .014}\\
whole pool & 0.1301 & $+0.061$ \ci{.054, .070} & 0.7202 & $-0.039$ \ci{$-$.046, $-$.031} & 0.8168 & $+0.017$ \ci{.010, .026}\\
\bottomrule
\end{tabular}}
\end{table}

\subsection{Held-out datasets}

On five held-out datasets (10 seeds; Figure~\ref{fig:main}b, Tables~\ref{tab:discovery} and~\ref{tab:heldout}), the discovered set is \emph{never significantly worse} than the plain network or the expert rule. On molbbbp it significantly beats both ($+0.035$ \ci{$+0.025,+0.045$} and $+0.016$ \ci{$+0.002,+0.030$} ROC-AUC). On the other four, point estimates lie in $[-0.012,-0.002]$ with every interval covering zero. The molbace interval against the expert rule only just covers zero (upper limit $+0.0003$).

\begin{table}[t]
\centering
\caption{Held-out datasets (frozen method, 10 seeds): every configuration, paired gain over the plain network with 95\% interval. Bold: validation-selected.}
\label{tab:heldout}
\footnotesize
\setlength{\tabcolsep}{3pt}
\resizebox{\linewidth}{!}{\begin{tabular}{@{}lccccc@{}}
\toprule
Config & molbace (AUC) & molbbbp (AUC) & PROTEINS (acc) & IMDB-B (acc) & REDDIT-B (acc)\\
\midrule
plain & 0.8084 & 0.6421 & 0.7786 & 0.7590 & 0.9175\\
expert rule & $-0.001$ \ci{$-$.019, .018} & $+0.019$ \ci{.011, .027} & $-0.009$ \ci{$-$.023, .005} & $-0.003$ \ci{$-$.016, .009} & $-0.001$ \ci{$-$.011, .011}\\
blk$_1$ & $-0.017$ \ci{$-$.039, .002} & $+0.019$ \ci{.010, .031} & $\mathbf{-0.003}$ \ci{$-$.015, .008} & $-0.019$ \ci{$-$.034, $-$.005} & $+0.008$ \ci{$-$.006, .022}\\
blk$_2$ & $-0.007$ \ci{$-$.027, .014} & $+0.032$ \ci{.016, .046} & $-0.013$ \ci{$-$.030, .001} & $-0.022$ \ci{$-$.041, $-$.002} & $\mathbf{-0.002}$ \ci{$-$.014, .012}\\
blk$_3$ & $\mathbf{-0.012}$ \ci{$-$.029, .008} & $\mathbf{+0.035}$ \ci{.025, .044} & $-0.003$ \ci{$-$.016, .011} & $\mathbf{-0.007}$ \ci{$-$.023, .010} & $-0.006$ \ci{$-$.017, .006}\\
random (size-matched) & $-0.024$ \ci{$-$.047, .001} & $+0.022$ \ci{.011, .033} & $-0.001$ \ci{$-$.015, .013} & $+0.001$ \ci{$-$.012, .015} & $+0.002$ \ci{$-$.009, .013}\\
whole pool & $-0.005$ \ci{$-$.026, .015} & $+0.047$ \ci{.037, .058} & $+0.004$ \ci{$-$.006, .016} & $+0.015$ \ci{.002, .027} & $-0.003$ \ci{$-$.013, .008}\\
\bottomrule
\end{tabular}}
\end{table}

\subsection{Out-of-fold versus in-sample evidence}
\label{sec:oof-evidence}

For the first pick, $\log_{10}p$ (scaled-$\chi^2$) out-of-fold vs.\ in-sample was $-9.1$ vs.\ $-3.9$ on molhiv. On the held-out datasets it was $-3.3$ vs.\ $-2.1$ (molbbbp), $-12.3$ vs.\ $-4.0$ (REDDIT-B), $-1.2$ vs.\ $-0.9$ (IMDB-B), $-4.1$ vs.\ $-4.6$ (molbace) and $-2.6$ vs.\ $-2.9$ (PROTEINS). The networks trained on half the data are weaker than those trained on all of it (ZINC MAE 0.279 vs.\ 0.191, molhiv ROC-AUC 0.737 vs.\ 0.759, COLLAB accuracy 0.758 vs.\ 0.800), which is the statistical debt of Remark~\ref{rem:debt}.

This weakness affects which rules are \emph{nominated}, not the gains we report. Every gain in Tables~\ref{tab:discovery}--\ref{tab:heldout} is measured on networks retrained on the full training set with the rule branch added, against the full-data plain network with the same seeds. A rule that only the half-data network fails to learn, and that a fully trained network absorbs, would therefore show no gain in these tables. On ZINC the ring term alone (blk$_1$) still cuts the full-data MAE from 0.1911 to 0.1181, so it is information that the fully trained network also leaves unlearned. Where a nominated block is worth little at full data (REDDIT-B: $\padj=0.001$, no gain), validation loss and the paired comparison expose it, which is why significance nominates and validation decides.

\subsection{In-distribution and shifted splits}

\begin{figure}[t]
\centering
\includegraphics[width=0.6\linewidth]{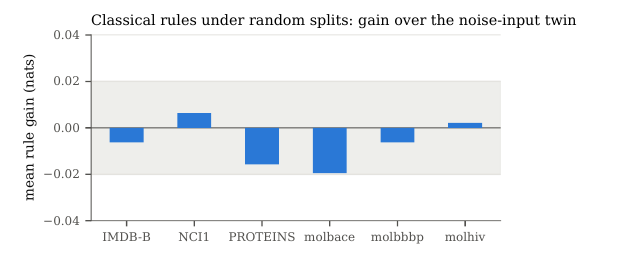}
\caption{Mean test log-loss gain of the classical rules over the noise-input twin under random splits. Every dataset lies within $\pm0.02$ nats (shaded band).}
\label{fig:shift}
\end{figure}

Under train-small/test-large splits, the plain networks are badly overconfident on large graphs: their test log-loss is 2.85 nats (NCI1) and 4.64 nats (PROTEINS), against 0.46 and 0.50 under random splits. Rule gains under size shift therefore mix information and calibration effects, which is why we report them without interpretation (Table~\ref{tab:shift}). Under random splits every classical rule set lies within $\pm0.02$ nats of its noise-input twin (Figure~\ref{fig:shift}).

\begin{table}[t]
\centering
\caption{Mean test log-loss gain of the classical rules over the noise-input twin (3 seeds), random split vs.\ shifted split (scaffold for molecules, train-small/test-large for TU). Plain: ROC-AUC (molecules) or accuracy (TU) of the plain network.}
\label{tab:shift}
\small
\begin{tabular}{@{}lcccc@{}}
\toprule
Dataset & plain (random) & plain (shifted) & mean rule gain, random & mean rule gain, shifted\\
\midrule
ogbg-molhiv & 0.828 & 0.756 & $+0.0021$ & $+0.0016$\\
ogbg-molbace & 0.862 & 0.810 & $-0.0195$ & $+0.0627$\\
ogbg-molbbbp & 0.954 & 0.664 & $-0.0062$ & $+0.0105$\\
NCI1 & 0.813 & 0.564 & $+0.0063$ & $+0.2112$\\
PROTEINS & 0.771 & 0.593 & $-0.0157$ & $+0.1702$\\
IMDB-B & 0.716 & 0.744 & $-0.0062$ & $-1.2528$\\
\bottomrule
\end{tabular}
\end{table}

\section{Scope}
\label{sec:scope}

Large gains concentrate where a structural term is unlearned (ring-dependent molecular targets); on social and protein graphs trained GINs leave little usable debt. Selection buys parsimony more than accuracy: the whole candidate pool sometimes matches the selected set (molbbbp $+0.047$ vs.\ $+0.035$) but can also hurt (molhiv $-0.039$). Significance nominates but does not measure size (REDDIT-B: $\padj=0.001$, no gain; Remark~\ref{rem:debt}). Under scaffold or size shift classical rules gain more on four of six datasets, but no in-distribution quantity predicted the shifted value. The Fano certificate is valid but loose (0.26--32 nats); the usable-information bound $\E[-\log q(S\mid T)]\ge\Hh(S\mid T)$ \citep{xu2020usable} is a tighter alternative. A discovered rule is a residual correlate of one trained network, not a causal mechanism; named chemotypes are hypotheses for domain experts.

\section{Conclusion}
\label{sec:conclusion}

A rule is worth what a GNN fails to learn, $\I(Y;S\mid T)$, of which expressivity controls only the first term. Out-of-fold residuals turn this identity into a calibrated search over named candidate rules, which recovered the hidden ring term of ZINC's target and transferred, frozen, to unseen datasets without significant loss and with a significant gain on BBBP.

\bibliographystyle{plainnat}
\bibliography{references}

\end{document}